\documentclass[runningheads]{llncs}
\usepackage{graphicx}

\usepackage{tikz}
\usepackage{comment} 
\usepackage{amsmath,amssymb} 
\usepackage{color}

\usepackage{dsfont}
\usepackage{siunitx}
\usepackage{mathtools}
\makeatletter
\newcommand{\printfnsymbol}[1]{%
  \textsuperscript{\@fnsymbol{#1}}%
}
\makeatother

\begin{document}
\pagestyle{headings}
\mainmatter
\def\ECCVSubNumber{1137}  

\title{InfoFocus: 3D Object Detection for Autonomous Driving with Dynamic Information Modeling} 

\titlerunning{InfoFocus}
%
\author{Jun Wang\inst{1}\thanks{Equal contribution.} \and
Shiyi Lan\inst{1}\printfnsymbol{1} \and
Mingfei Gao\inst{2} \and
 Larry S. Davis\inst{1} }
\authorrunning{J. Wang, S. Lan et al.}
%
\institute{$^1$University of Maryland, College Park MD 20742, USA \\
$^2$Salesforce Research, Palo Alto, CA 94301, USA \\
\email{\{junwang,lsd\}@umiacs.umd.edu, sylan@cs.umd.edu}\\
\email{mingfei.gao@salesforce.com}}
\maketitle

\begin{abstract}
Real-time 3D object detection is crucial for autonomous cars. Achieving promising performance with high efficiency, voxel-based approaches have received considerable attention. However, previous methods model the input space with features extracted from equally divided sub-regions without considering that point cloud is generally non-uniformly distributed over the space. To address this issue, we propose a novel 3D object detection framework with dynamic information modeling. The proposed framework is designed in a coarse-to-fine manner. Coarse predictions are generated in the first stage via a voxel-based region proposal network. 
We introduce InfoFocus, which improves the coarse detections by adaptively refining features guided by the information of point cloud density. Experiments are conducted on the large-scale nuScenes 3D detection benchmark. Results show that our framework achieves the state-of-the-art performance with 31 FPS and improves our baseline significantly by 9.0\% mAP on the nuScenes test set.

\end{abstract}

\begin{figure}[t]
\centering
  \includegraphics[width=1.0\linewidth]{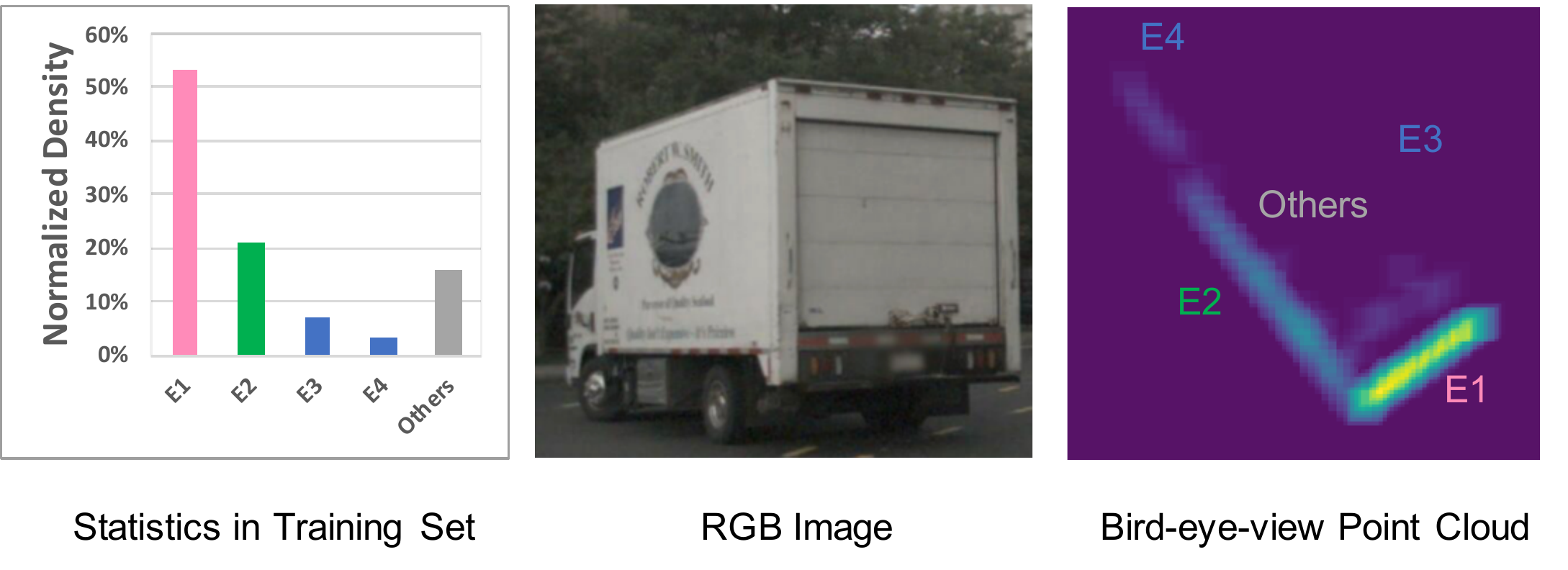}
  \\
  \caption{Left: we calculate the average point density across different parts of objects in BEV of nuScenes training set. \emph{E1}-\emph{E4} indicate four edges sorted by their normalized density scores (sum to 100\%) and \emph{others} denotes areas inside objects. We set each edge width as 10\% of the length along the object size and only objects over more than 100 points are counted. Middle and right: we visualize an example of the LiDAR point cloud in 2D image and its corresponding  bird's eye view (BEV). Clearly, most of the point clouds locate on the contour of the object}
\label{fig:intro}
\end{figure}

\section{Introduction}
With growing interests in autonomous vehicles, 3D object detection has received considerable attention. Due to the superior capability of modeling 3D objects, point cloud is the most popular type of data source. Most existing 3D detectors are point-based~\cite{qi2018frustum,wang2019frustum,Lan_2019_CVPR,shi2019pointrcnn,yang2019std} and voxel-based~\cite{lang2019PointPillars,zhou2018voxelnet,yan2018second,ye2020sarpnet,hu2019you}. Point-based approaches generate features from raw point cloud data directly. Although achieving promising performance, these methods suffer from high computational complexity which discourages their application in real-time scenarios. Voxel-based approaches~\cite{lang2019PointPillars,zhou2018voxelnet,yan2018second,ye2020sarpnet,hu2019you} firstly convert point cloud into voxels and then employ deep convolutional neural networks (DCNN) to conduct object detection. Taking advantage of the advanced DCNN architecture, voxel-based approaches achieve the state-of-the-art performance with low computational cost. Our work follows the setting of voxel-based methods for their advanced balance of efficiency and effectiveness.

Although much progress has been made in improving the performance of voxel-based detectors, an important characteristic of point cloud is not well explored: input data points are usually not uniformly distributed over the space. The density of point cloud can be affected by different factors, e.g., the distance of objects from LiDAR sensor and object self-occlusion. As illustrated in Fig.~\ref{fig:intro}, the density of point cloud over objects highly depends on the relative locations of different parts. It is also intuitive that the amount of information is highly related to the point density. However, existing voxel-based detectors extract features from uniformly divided sub-regions, regardless of the actual distribution of the points. We believe that this will lead to loss of useful information and ultimately result in sub-optimal detection performance. 

To fully exploit the non-uniform distribution of point cloud, we propose a novel 3D object detection framework, to adaptively model the rich feature of 3D objects according to the information density of points. Illustrated in Fig.~\ref{fig:arch}, our framework contains two stages. Coarse detection results are obtained in the first stage via voxel-based region proposal network. In the second stage, we introduce InfoFocus, to model and extract the informative features from regions of interest (RoI) (formed by the coarse predictions) according to the distribution of point cloud, and the predictions are improved with the help of the refined features.

The InfoFocus is the core structure of our framework which contains three sequentially connected modules including the Point-of-interest (PoI) Pooling, the Visibility Attentive Module, and the Adaptive Point-wise Attention. 

\textbf{PoI Pooling.} Unlike 2D objects which contain densely distributed information over the whole RoI, more of the points of 3D objects locate on the their surfaces. Therefore, we hypothesize that most informative feature concentrates on the edge of RoI. Motivated by this intuition, we propose PoI Pooling which densely samples features on the edge and sparsely samples feature in the middle of RoI to accommodate the non-uniform information distribution of point cloud.

\textbf{Visibility Attentive Module.} Heavy self-occlusion is presented because of the nature of LiDAR data that is no point cloud exists on the backside of object relatively to the sensor. To mitigate this issue, our proposed Visibility Attentive Module applies hard attention to emphasize the visible parts of objects and eliminate the features from invisible points.

\textbf{Adaptive Point-wise Attention.} PoIs may contain different amount of information, although they are all visible. We introduce Adaptive Point-wise Attention to re-weight the features to improve the modeling of 3D objects.

We conduct extensive experiments on the largest public 3D object detection benchmark, i.e, nuScenes~\cite{caesar2019nuscenes}. Experimental results show that our approach significantly outperforms the baselines, achieving $39.5\%$ mAP  with $31$ FPS. Results of comprehensive ablation studies demonstrate the effectiveness of our InfoFocus and that each sub-module makes considerable contributions to our framework.

\section{Related Work}

\noindent\textbf{Point-based Detectors}.
Inspired by the powerful feature learning capability of PointNet \cite{qi2017pointnet,qi2017pointnet++} and the advanced modeling structure of 2D object detectors \cite{Girshick_2014_CVPR,Girshick_2015_ICCV,ren2015faster}, Frustum PointNets~\cite{qi2018frustum} extrude the 2D object proposals into frustums to generate the 3D bounding boxes from raw point cloud. Lan et al.~\cite{Lan_2019_CVPR} add a decomposition-aggregation module modeling local geometry to extract the global feature descriptor of point cloud. Limited by initial 2D box proposals, those methods yield low performance when objects are occluded. In contrast, PointRCNN \cite{shi2019pointrcnn} generates 3D proposals directly from point cloud instead of 2D images. The recent STD \cite{yang2019std} attempts to refine the detection boxes in a coarse-to-fine manner. However, all those methods are computationally expensive due to the large amount of data points to be processed. 

\noindent\textbf{Multi-view 3D Detectors}.
MV3D~\cite{chen2017multi} is proposed to fuse multi-view feature maps for the generation of 3D box proposals. Following \cite{chen2017multi}, Ku et al. \cite{ku2018joint} explore high resolution feature maps to compensate the information loss for small objects. These methods address the feature alignment between multi-modality in a coarse level and are typically slow. Liang et al.~\cite{Liang_2018_ECCV} design a continuous fusion layer to deal with the continuous state of LiDAR and the discrete state of images. Later, ~\cite{liang2019multi,vora2019pointpainting} leverage different strategies to jointly fuse related tasks to improve feature representation.

\noindent\textbf{Voxel-based Detectors}.
Recently, there is a trend of using regular 3D voxel grids to represent point cloud such that the input data can be easily processed by the 3D/2D convolution networks. Among those, VoxelNet~\cite{zhou2018voxelnet} is the pioneering work of performing voxelization on the raw 3D point cloud. To improve its efficiency, Second~\cite{yan2018second} adopts Sparse Convolution and speeds up detection process without compromising the detection accuracy. PointPillars~\cite{lang2019PointPillars} dynamically converts the 3D point cloud into a 2D pseudo image, making it more suitable for the application of the existing 2D object detection techniques. In ~\cite{ye2020sarpnet}, Ye et al. design a new voxel generator to preserve the information loss along the vertical direction. Building upon voxel-based detectors, our model captures richer information of objects by refining their feature representations at a second stage guided by the point cloud density and ultimately improves the detection results. 

There are several recent studies~\cite{chen2019fast,liu2019point} focusing on fusing the voxel-based features with PointNet-based features in order to extract more fine-grained 3D features. InfoFocus is complementary to these techniques and can be further applied on top of them. WYSIWYG~\cite{hu2019you} is the most related method to our approach since we both drive the model to encode visibility information. However, instead of using a separate branch to generate the hidden invisibility representation, our method directly aggregates the valuable point-wise features together from existing backbone network to refine the proposals in an end-to-end manner. 

\section{Proposed Approach}
\begin{figure}
\centering
  \includegraphics[width=1.0\linewidth]{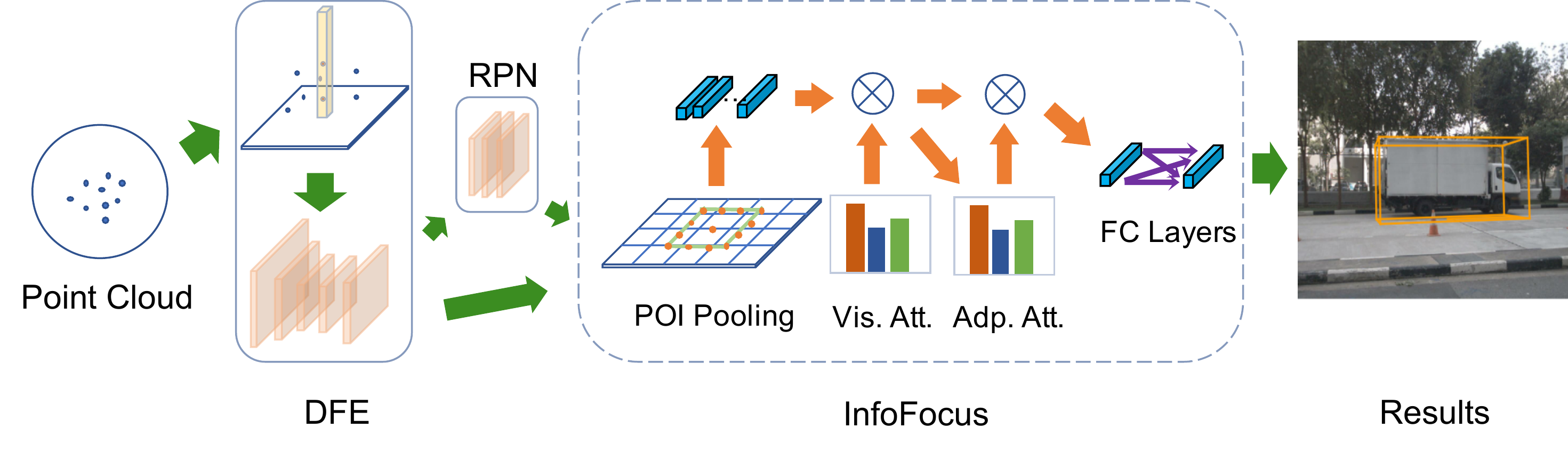}
  \\
  \caption{The proposed 3D object detection framework. It consists of three parts: Deep Feature Extractor(DFE), Region Proposal Network, and InfoFocus. InfoFocus contains three modules: PoI Pooling, Visibility Attentive Module, and Adaptive Point-wise Attention Module}
\label{fig:arch}
\end{figure}

The proposed framework is illustrated in Fig.~\ref{fig:arch}, which consists of a deep feature extractor followed by a two-stage architecture. The deep feature extractor containing a Pillar Feature Network and a DCNN, converts the input point cloud to representative feature maps. Specifically, the Pillar Feature Network divides the whole space into equal pillars and generates the so-called pseudo images~\cite{lang2019PointPillars}. The pseudo images are then processed by the DCNN to obtain the feature maps which are shared by the two stages, i.e., Region Proposal Network (RPN) and InfoFocus. The RPN generates the initial coarse bounding box proposals that are refined by InfoFocus, with dynamic information modeling. Note that our Deep Feature Extractor and RPN follow the setting of~\cite{lang2019PointPillars}.

\subsection{Deep Feature Extractor}
Deep Feature Extractor is composed of two parts: 1) voxelization using Pillar Feature Network that converts the orderless point cloud into a sparse pseudo image via a simplified PointNet-like architecture and 2) feature extraction using DCNN to learn informative feature maps.

\noindent\textbf{Pillar Feature Network}.
The Pillar Feature Network operates on the raw point cloud, and learns point-wise features for each pillar. After voxelizing raw point cloud into evenly spaced pillars, we randomly sample $N$ points from each non-empty pillar and then obtain a dense tensor with the size of $ D \times P \times N $, where D indicates the information dimension of each point, P denotes the number of non-empty pillars, and N denotes the number of points in each pillar. The Pillar Feature Network utilizes a PointNet-like block to learn a multi-dimensional feature vector for each pillar. The pillar-wise features are encoded into a 2D pseudo image with the shape of $ W \times L \times C $, where $W$ and $L$ indicate the width and length of the pseudo image, and $C$ is the channel of the feature map. 

\noindent\textbf{Deep Convolution Neural Network (DCNN)}.
DCNN learns feature maps from the generated pseudo 2D image. The DCNN uses conv-deconv layers to extract features of different levels, and concatenates them to get the final features from different strides.

\subsection{Region Proposal Network (RPN)}
The RPN takes the feature maps provided by DCNN as inputs and produces high-quality 3D object proposals. Similar to the proposal generation in 2D object detection, anchor boxes are predefined at each position and proposals are generated by learning the offsets between anchors and the ground truths. To handle different scales of objects, a dual-head strategy is adopted. Specifically, the small-scale head takes features from the first conv-deconv phase of the DCNN, while the large-scale head takes the features from its concatenation phase.

\subsection{InfoFocus}
The InfoFocus serves as the second stage of our framework, which takes the candidate proposals from RPN and extracts features of objects in a hierarchical manner from the feature maps produced by the DCNN. Specifically, given each 3D object proposal, InfoFocus dynamically focuses on the informative parts of the feature maps by gradually emphasizing the representative PoIs in the following three steps: 1) the edge points are selected out from the whole proposal region by PoI Pooling; 2) Visibility Attention module emphasizes on the informative points according to their relative visibility to the LiDAR sensor and 3) in the Adaptive Point-wise Attention module, the features of the visible points are further weighted adaptively. The re-weighted features of the visible points are then fused to form the final representation of the proposal, on top of which two fully-connected layers are utilized to predict the refined box.

\begin{figure}[t]
\centering
  \includegraphics[width=0.7\linewidth]{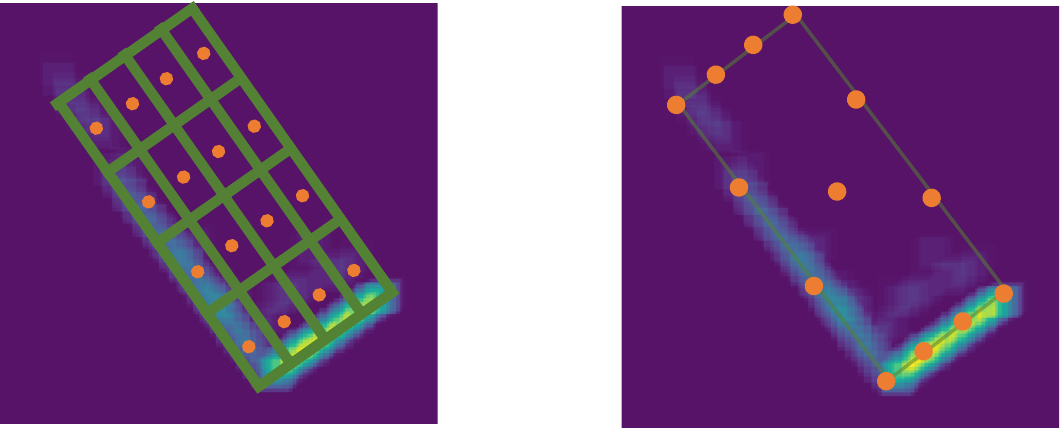}
  \caption{RoI Pooling  \emph{vs.} PoI Pooling. The grid represents the feature map, and the dots denotes sampling points of interest. RoI Pooling samples the whole box, while PoI Pooling focuses on the key-points from edge-of-interest}
\label{fig:POIPooling}
\end{figure}

\noindent\textbf{PoI Pooling}.
When representing a 3D proposal, the most intuitive way is adopting the commonly used strategy in the two-stage 2D object detectors, i.e., RoI Pooling (see Fig.~\ref{fig:POIPooling} left). However, unlike the 2D images that have densely distributed information over the region proposals, the 3D point cloud mostly resides on the object surface which results in non-uniform information over the regions (most information locates on the edges of proposals).

The proposed PoI pooling is illustrated in Fig. ~\ref{fig:POIPooling} (right). Instead of equally sampling points over a region of the feature maps, we focus on sampling the points on the informative parts including four corners, the center point and key-points on the edges. Note that we consider the center position as an additional useful signal since it is likely to capture the semantic-level information. 

We first project the 3D proposal to the birds' view coordinate system. Let $p_{0}, p_1, p_2$ and $p_3$ represent the positions of top-left, top-right, bottom-right, and bottom-left corners of a proposal on the pseudo image, respectively and $p_c$ denotes the center point. Along each edge, $n$ more key-points are uniformly sampled. For example, for the top edge between $p_0$ and $p_1$, the position of a sampled key-point $kp_j= (p_{0}\frac{j}{n+1} + p_{1}\frac{n+1-j}{n+1})$, where $j$ is an integer and $ 1 \leq j \leq n$. 
To this end, $(5 + 4*n)$ PoIs are obtained. A high-dimensional feature is extracted for each PoI according to its relative position on the feature map and then we obtain a feature set $F_{poi}=\{f^{poi}_1, f^{poi}_2,...,f^{poi}_{N_{poi}}\}$, where $N_{poi}=(5 + 4*n)$ representing the number of selected PoIs within the considered region. 

\begin{figure}[t]
\centering
  \includegraphics[width=1.0\linewidth]{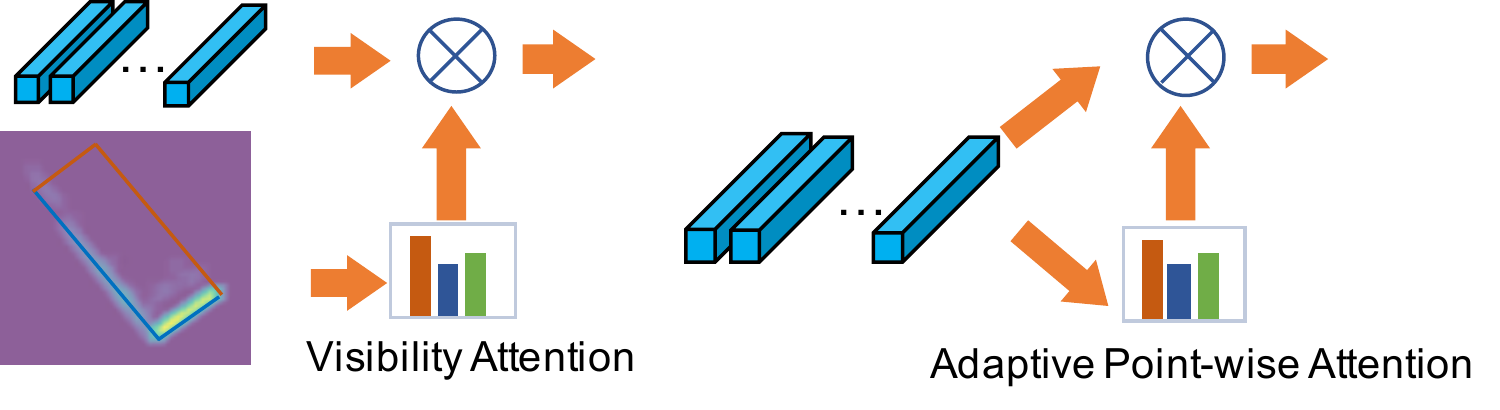}
  \\
  \caption{Left: illustration of the Visibility Attentive Module. We compute hard attention for each sampled point depending on whether it is visible to the sensor. We also show the visibility map on the bottom left. Points on the blue line are visible while points on the orange line are invisible. Right: the architecture of the Adaptive Point-wise Attention Module. The point-wise attention is generated using a fully connected (FC) layer followed by a Sigmoid function. The input of FC layer is the feature of each point}
\label{fig:visibility}
\end{figure}

\noindent\textbf{Visibility Attentive Module}.
Severe self-occlusion typically occurs in point cloud, but is ignored by most of the existing methods. The Visibility Attentive Module (see Fig. \ref{fig:visibility} left) is proposed to mitigate this issue by focusing on the information provided by the visible parts of objects. We argue that visible regions contain more useful information than the occluded ones. Formally, we propose to re-weight features of PoIs according to their corresponding visibility by exploiting the geometric relationship between the proposals and the LiDAR sensor in bird’s eye view. As shown in Eq.~\ref{eq: vam}, $F_{vis}$ denotes the updated feature set, where $v^{poi}_i$ indicates the visibility score of the $i$th PoI. Different weighting strategies can be used and we use a hard attention strategy in this work for its simplicity, that is assigning $v^{poi}_i=1$ if the $i$th PoI is visible and $v^{poi}_i=0$ otherwise. In other words, we only take PoIs on the visible edges to represent the proposal. 

\begin{equation}
\label{eq: vam}
F_{vis} = \{ f^{poi}_1 * v^{poi}_1, f^{poi}_2 * v^{poi}_2,..., f^{poi}_{N_{poi}} * v^{poi}_{N_{poi}}\}
\end{equation}

For the consideration of model efficiency, a simple yet effective method is used to estimate the visibility of points in the bird's eye view. To figure out the sides of proposals facing to the sensor, we first compute the distance of each corner to the LiDAR sensor and determine the one that is closest to the sensor. Then, we consider the two edges passing this closest corner as the visible edges and the other two as the occluded ones. 

\noindent\textbf{Adaptive Point-wise Attention Module}.
PoI Pooling and Visibility Attentive Module are motivated by the nature of the non-uniform density of point cloud. However, two points may offer different amount of information even though they are all visible by the sensor. Adaptive Point-wise Attention Module provides the flexibility for the visible PoIs to contribute unequally to the prediction. Suppose $F_{vis} = \{ f^{vis}_{1}, f^{vis}_{2}, ..., f^{vis}_{N_{vis}} \}$ indicates the feature set of visible PoIs. Adaptive Point-wise Attention Module learns an attention weight, $w_i$, for each $f^{vis}_i$ adaptively for the next-step feature aggregation. Specifically, a shared fully connected (FC) layer with sigmoid as the activation function is used to learn the attention weights, formally expressed as $v^{vis}_i = Sigmoid(\textbf{W}f^{vis}_i+b)$. We use $F_{att}=\{ f^{att}_{1}, f^{att}_{2}, ..., f^{att}_{N_{vis}} \}$ to represent the re-weighted feature set of visible PoIs updated using $F_{vis}$ and the attention weights, where $f^{att}_i=f^{vis}_i*v^{vis}_i$.
 
The final representation of each proposal aggregates the features of its visible PoIs. Let $e_0, e_1, e_2$ and $e_3$ denote the top, right, down, left edges of a proposal, respectively. We first compute $f^e_i$ by applying max pooling to all the visible points on $e_i$. Then, the final representation is obtained by $f^e_0||f^e_1||f^e_2||f^e_3||f^p_c$, where $f^p_c$ indicates the feature of the center point and $||$ indicates concatenation.

\subsection{Loss Function}
Given the output PoI feature representation from the aforementioned three modules topped by fully-connected layers, the head network consists of three branches predicting the box class, localization and direction. The ground truth and anchor boxes are parameterized as $ (x, y, z, w, l, h, \theta) $, where $(x, y, z)$ is the center of box, $(w, l , h)$ is the dimension of box, and $\theta$ is the heading along the z-axis in the LiDAR coordinate system. The box regression target is computed as the residuals between the ground truth and the anchors as:

\begin{equation}
\begin{split}
    \triangle x = \frac{x^{gt} - x^a}{d^a},  \triangle y &= \frac{y^{gt} - y^a}{d^a},  
\triangle z = \frac{z^{gt} - z^a}{h^a}, \\
\triangle w = \log (\frac{w^{gt}}{{w^{a}}}),  \triangle l &= \log (\frac{l^{gt}}{{l^{a}}}),  
\triangle h = \log (\frac{h^{gt}}{{h^{a}}}), \\
\triangle \theta &= \theta_{gt} - \theta_{a}
\end{split}
\end{equation} 

where $x^{gt}$ and $x^a$ refer to ground truth and anchor box respectively, and $ d^a = \sqrt{(w^a)^2 + (l^a)^2} $. To deal with severe class imbalance problem in the dataset, we adopt the focal loss \cite{lin2017focal} for the classification loss. Smooth L1 loss \cite{Girshick_2014_CVPR} is used for the regression loss. In addition, to compensate for direction prediction missing in the regression, we adopt a softmax classification loss on orientation prediction. 

Similar with that of the vanilla PointPillars network \cite{lang2019PointPillars}, we formally define a multi-task loss for both stages as threefold, 

\begin{equation}
 L_{stage\_i} = \frac{1}{N_{pos}} (\beta_{cls} L_{cls\_i} + \beta_{reg} L_{reg\_i} + \beta_{dir} L_{dir\_i} ),
\end{equation}

where i could be either RPN or InfoFocus stage, $N_{pos}$ refers to the number of positive anchors and $  \beta_{cls}, \beta_{reg}, \beta_{dir} $ are chosen to balance the weights among classification loss, regression loss and direction loss. 

\subsection{Comparisons with Existing Approaches}
\noindent\textbf{Point-based Approaches}. Our framework uses PointNet to extract features from equally divided sub-grids and employs a DCNN to generate 2D feature maps while point-based techniques \cite{qi2018frustum,wang2019frustum,Lan_2019_CVPR,shi2019pointrcnn} only use PointNet as its backbone. Both our approach and point-based approaches apply two-stage architecture to infer objects. Meanwhile, we both sample features considering the distribution of point cloud. However, compared to PointNet, InfoFocus is more computationally efficient without performance degradation.

\noindent\textbf{Fusion-based Approaches}. Fusion-based detectors \cite{chen2019fast,liu2019point} make use of both RGB images and point cloud data for 3D object detection. InfoFocus is much faster than fusion-based approaches, since they contain two backbones to process multi-view sources and are heavily engineered. On the other hand, InfoFocus also achieves competitive results compared to fusion-based approaches. 

\noindent\textbf{Traditional Voxel-based Approaches}. Our method shares the similar backbone as the existing voxel-based architectures \cite{lang2019PointPillars,zhou2018voxelnet,yan2018second,ye2020sarpnet}. However, previous voxel-based detectors pay less attention to the distribution of LiDAR data that most 3D point cloud locates on the surface of the objects. Our proposed PoI Pooling, Visibility Attentive Module, and Adaptive Point-wise Attention model the non-uniform point cloud using dynamic information focus. First, the PoI Pooling decreases the sampling from the inside of objects where few points locate. Next, the Visibility Attentive Module eliminates the noise from the back of objects where points are occluded. Last, we apply the Adaptive Point-wise Attention to learn the focus on each sampled points. Jointly, these modules contribute significantly to the superior performance of InfoFocus.

\section{Experiments}
Our method is mainly evaluated on the nuScenes dataset~\cite{caesar2019nuscenes} which is considered as the most challenging 3D object detection benchmark. We first present our implementation details. We compare with the existing approaches both quantitatively and qualitatively. Then, extensive ablation studies are conducted to demonstrate the effectiveness of each designed module. Last, we analyze the inference time and the desired speed accuracy trade-off provided by our method.

\subsection{Dataset and Evaluation Metric}
NuScenes~\cite{caesar2019nuscenes} is one of the largest datasets for autonomous driving. There are 1000 scenes of 20$s$ duration each, including 23 object classes annotated with 28,130 training, and 6,019 validation samples. We use the LiDAR point cloud as the only input to our method and all the experiments follow the standard protocol on the training and validation sets. Officially, nuScenes evaluates the detection accuracy across different classes, based on the average precision metric (AP) which is computed based on 2D center distance between ground truth and the detection box on the ground plane. In details, the AP score is determined as the normalized area under the precision recall curve above 10\%. The final mean AP (mAP) is the average among the set of ten classes over matching thresholds of $\mathds{D} = \{0.5, 1, 2, 4 \} $ meters.

\subsection{Implementation Details}
We integrate InfoFocus into a state-of-the-art real-time 3D object detector~\cite{lang2019PointPillars} to improve the detection performance without largely compromising speed. Closely following the codebase\footnote{https://github.com/traveller59/second.pytorch.} recommended by the authors of PointPillars~\cite{lang2019PointPillars}, we use PyTorch to implement our InfoFocus modules and integrate it into vanilla PointPillars network. More details will be introduced in the supplementary materials. 

\noindent\textbf{RPN}. For each class of objects, the RPN anchor size is set by calculating the average of all objects from the corresponding class in training set of nuScenes. In addition, the matching thresholds are based on the custom configuration following the suggested codebase. 1,000 proposals are obtained from RPN, on which NMS with a threshold of 0.5 is applied to remove the overlapping proposals for both training and inference. The final top-ranked 300 proposals are kept for the InfoFocus stage to simultaneously predict the category, location and direction of objects during both the training and inference.

\noindent\textbf{InfoFocus}. The second stage is our proposed InfoFocus. The three novel modules process object-centric feature sequentially based on the initial bounding box proposals from RPN. The number of sampled key-points for each edge, $n$, is set to be 2. Thus, the total number of PoIs, $N_{poi}$, is 13, including a center, 4 corners and 2 key-points on each edge. Similar to RoIAlign~\cite{he2017mask}, bi-linear interpolation is used to compute the deep feature from four neighboring regular locations of each point.

As mentioned before, we apply a max-pool layer to summarize the features of points along each edge, resulting in 5 features for each proposal, including features from top, right, down, left edges and the center. When concatenating these features, we always treat the edge that is closest to the sensor as the top edge. A fully connected layer with a single node is used to generate point-wise attention weight for each point.

The feature of each proposal is transformed by two consecutive FC layers with 512 nodes each and passed to three sibling linear layers, a box-regression branch, a box-classification branch and a box-direction branch. For the regression target assignment, anchors having Intersection over Union (IoU) bigger than 0.6 with the ground truth are considered positive, and smaller than 0.55 are assigned negative labels. 

\noindent\textbf{Training Parameters}.
Experiments are conducted on a single NVIDIA 1080Ti GPU. The weight decay is set to be 0.01. We adopt the Adam optimizer \cite{kingma2014adam}, and use a one-cycle scheduler proposed in \cite{smith2018disciplined}. We train our model with a total of 20 epochs as a default choice, taking about 40 hours from scratch. For the first 8 epochs, the learning rate progressively increases from \num{3e-4} to \num{3e-3} with decreasing momentum from 0.95 to 0.85, while in the remaining 12 epochs learning rate decreases from \num{3e-3} to \num{3e-6} with increasing momentum from 0.85 to 0.95. The focal loss \cite{lin2017focal} with $ \alpha = 0.25 $ and $ \gamma = 2.0 $  is adopted for the classification loss. The balancing weights for the classification, box regression, and direction loss \(  \beta_{cls}, \beta_{reg}, \beta_{dir} \) of both stages are 1, 2 and 0.2, respectively.

\subsection{Main Results}
\begin{figure}[t]
\centering
  \includegraphics[width=1.0\linewidth]{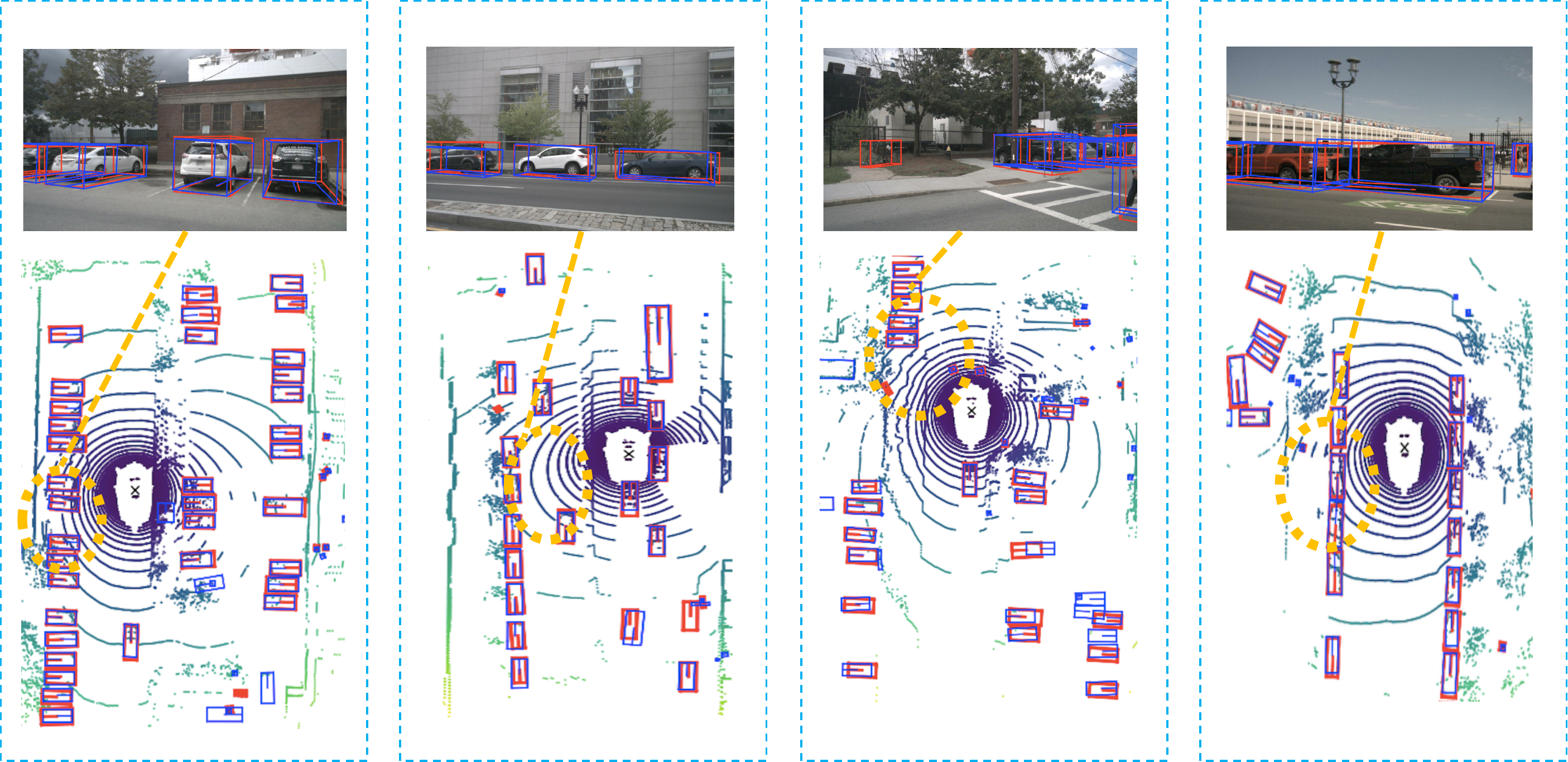}
  \\
  \caption{We visualize the detection results on nuScenes with 2D and 3D BEV images. On the top, we demonstrate the 2D images with the 3D bounding box annotated, while the BEV of LiDAR with ground truth (red) and detection (blue) box are shown on the bottom. Note that the line in the frame denotes the direction of the object}
\label{fig:visualization_2D_3D}
\end{figure}

\begin{figure}[t]
\centering
  \includegraphics[width=1.0\linewidth]{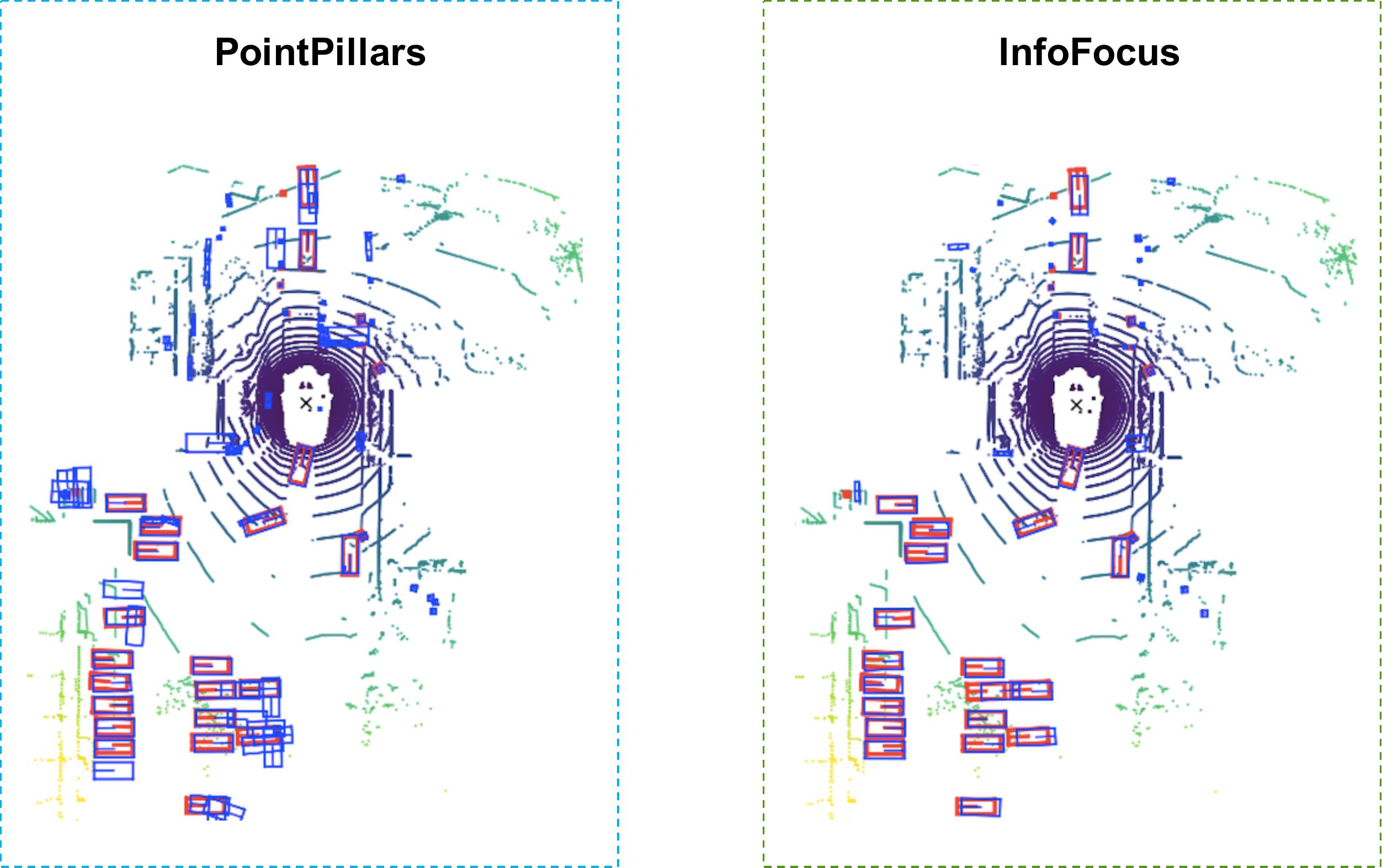}
  \\
  \caption{We visualize the BEV detection results for the same point cloud sample on nuScenes with the vanillar PointPillars (left) and InfoFocus (right)}
\label{fig:comparison}
\end{figure}

 First, we compare our framework with the state-of-the-art methods on the nuScenes validation set, including the vanilla PointPillars~\cite{lang2019PointPillars} as our baseline, and recently published WYSIWYG~\cite{hu2019you}. As can be seen from Table.~\ref{table:1}, the baseline has an mAP of 29.5\% with a single stage, while InfoFocus improves it by a massive 6.9\%. This demonstrates the effectiveness of InfoFocus. We also visualize the detection results of our framework on 2D and 3D BEV images in Fig.~\ref{fig:visualization_2D_3D}. As shown in Fig.~\ref{fig:comparison}, compared to the vanilla PointPillars qualitatively, InfoFocus helps remove the false positives significantly and obtains better results. 
 
  \begin{table}[h!]
\centering
\caption{Object detection results (\%) on nuScenes validation set}
\begin{tabular}{l c c c c c c c c c c c} 
 \hline
 Method & car & peds. & barri. & traff. & truck & bus & trail. & const. & motor. & bicyc. & mAP \\ [0.5ex] 
 \hline
  PointPillars \cite{lang2019PointPillars} & 70.5 & 59.9 & 33.2 & 29.6 & 25.0 & 34.3 & 16.7 & 4.5 & 20.0 & 1.6 & 29.5 \\

    WYSIWYG \cite{hu2019you} & 80.0 & 66.9 & 34.5 & 27.9 & 35.8 & 54.1 & 28.5 & 7.5 & 18.5 & 0 & 35.4 \\
  \hline
  Ours & 77.6 & 61.7 & 43.4 & 33.4 & 35.4 & 50.5 & 25.6 & 8.3 & 25.2 & 2.5 & \bf 36.4 \\
\hline
\end{tabular}
\label{table:1}
\end{table}

 In addition, we submit the detection results of test set on the nuScenes test server. The results show that our method achieves the state-of-the-art performance with inference speed of 31 FPS, improving the baseline performance by 7\% mAP. Note that all methods listed in Table.~\ref{table:2} are LiDAR-based except that MonoDIS \cite{simonelli2019disentangling} and CenterNet\cite{zhou2019objects} are camera-based methods. Without bells and whilstles, our approach works better than WYSIWYG~\cite{hu2019you}. 
 Considering that our model contains more parameters than the vanilla PointPillars, we empirically increase the number of the training epoch by 2 times. With all the others settings the same, our method is improved by 2\% mAP on the nuScenes test set as shown in Table.~\ref{table:2} (Ours 2$\times$). In total, our method outperforms WYSIWYG\cite{hu2019you} by 4.5\% mAP on the nuScenes test set. In the rest of paper, the default setting of training epochs is adopted.
 To the best of our knowledge, our framework is superior than all the published real-time methods with respect to mAP.

\begin{table}[h!]
\centering
\caption{Object detection results (\%) on nuScenes test set. Note that MonoDIS and CenterNet are camera based methods, and the rest are LiDAR based. Ours 2$\times$ indicates 2$\times$ training time with other settings being the same with Ours}
\begin{tabular}{l c c c c c c c c c c c} 
 \hline
 Method & car & peds. & barri. & traff. & truck & bus & trail. & const. & motor. & bicyc. & mAP \\ [0.5ex] 
 \hline
  MonoDIS \cite{simonelli2019disentangling} & 47.8 & 37.0 & 51.1 & 48.7 & 22.0 & 18.8 & 17.6 & 7.4 & 29.0 & 24.5 & 30.4 \\
  PointPillars \cite{lang2019PointPillars}  & 68.4 & 59.7 & 38.9 & 30.8 & 23.0 & 28.2 & 23.4 & 4.1 & 27.4 & 1.1 & 30.5\\
  SARPNET \cite{ye2020sarpnet} & 59.9 & 69.4 & 38.3 & 44.6 & 18.7 & 19.4 & 18.0 & 11.6 & 29.8 & 14.2 & 32.4 \\
  CenterNet \cite{zhou2019objects} & 53.6 & 37.5 & 53.3 & 58.3 & 27.0 & 24.8 & 25.1 & 8.6 & 29.1 & 20.7 & 33.8 \\ 
  WYSIWYG \cite{hu2019you}  & 79.1 & 65.0 & 34.7 & 28.8 & 30.4 & 46.6 & 40.1 & 7.1 & 18.2 & 0.1 & 35.0 \\

  \hline
   Ours & 77.2 & 61.5 & 45.3 & 40.4 & 31.5 & 44.1 & 35.9 & 9.8 & 25.1 & 4.0 & \bf 37.5 \\ 
   Ours 2$\times$ & 77.9 & 63.4 & 47.8 & 46.5 & 31.4 & 44.8 & 37.3 & 10.7 & 29.0 & 6.1 & \bf 39.5 \\ 

 \hline
\end{tabular}
\label{table:2}
\end{table}

\subsection{Ablation Studies}
To understand the contribution of our major component to the success of InfoFocus, Table.~\ref{table:3} summarizes the performance of our framework when a certain module is disabled, including PoI Pooling, Visibility Attention Module and Adaptive Attention Module.

\begin{table}[]
\centering
\caption{Ablation studies on nuScenes validation set. "Vis. Att" and "Adp. Att." refer to Visibility Attention Module and Adaptive Attention Module, respectively}
\label{table:3}
\medskip    
\begin{tabular}{c@{\ \ \ \ \ \ } c@{\ \ \ \ \ \ \ } c@{\ \ \ \ \ \ \ }@{\ \ \ } c}
\hline
PoIPool & Vis. Att. & Adp. Att. & mAP   \\ 
 \hline
&    &    & 29.5  \\ 
  \checkmark &     &     &  32.5    \\ 
\checkmark & \checkmark &           & 34.8    \\ 
\checkmark &        & \checkmark  &    34.8  \\ 
    \checkmark  & \checkmark & \checkmark & 36.4 \\  
\hline
\end{tabular}
\end{table}

\noindent\textbf{PoI Pooling}.
To investigate the effect of PoI Pooling, we simply add the PoI Pooling on top of the vanilla PointPillar. This baseline introduces 3.0\% mAP improvement. However, when we vary the number of pooling key-points on each edge, we see that our framework with four key-points ($n=4$) on each edge degrades slightly by 0.8\% mAP than that of two key-points ($n=2$). A possible reason is that the higher number of samples along each edge might bring more noise which harms the detection performance. 

\noindent\textbf{Visibility Attention}.
We further add the Visibility Attention module to filter out invisible edges before PoI pooling. Table.~\ref{table:3} shows that when using the features from two visible edges, the mAP result is improved  by 2.3\% mAP compared to \emph{baseline+PoIPool}. Generally, the visible parts of objects correspond to their sides closer to the LiDAR sensor, thus they may capture richer information. By applying visibility attention, our method focuses more on the representative information which results in better performance.

\noindent\textbf{Adaptive Point-wise Attention}.
Without the Adaptive Point-wise Attention module, the framework naturally allows the same weight for each PoI feature. As we can see in Table.~\ref{table:3}, when adding this module, the result of \emph{baseline+PoIPool} improves by 2.3\% mAP and that of \emph{baseline+PoIPool+Vis.Att.} improves by 1.6\%. These results suggest that the Adaptive Point-wise Attention module helps emphasize on useful points which leads to a better performance.

\begin{table}[]
\centering
\caption{Inference time of 3D object detectors. Note that inference time for the baseline here is the network reproduced by ourselves}
\label{table:4}
\medskip

\begin{tabular}{l@{\ \ \ \ \ } c@{\ \ \ \ \ }  c@{\ \ \ \ \ } c@{\ \ \ \ \ }} 
 \hline
 Method & Input Format & mAP &  Inference Time (ms) \\[0.5ex]
 \hline
Baseline \cite{lang2019PointPillars} & LiDAR & 30.5 & 26.9 \\
MonoDIS \cite{simonelli2019disentangling} & RGB & 30.4 & 29.0  \\
SARPNET \cite{ye2020sarpnet} & LiDAR & 32.4& 70.0  \\
\hline
Ours & LiDAR & 37.5 & 32.9 \\

\hline
\end{tabular}

\end{table}

\noindent\textbf{Rotated RoIAlign Comparison}.
One widely considered way to extract the region-wise features in the two-stage architecture is RoIAlign~\cite{he2017mask}. So, it is intuitive to compare with this strategy under the setting of 3D object detection. We implement rotated RoIAlign (RRoI) operation \cite{huang2018improving} to compensate for the rotated bounding box, since in our case they are often not axis-aligned. We conduct experiments exploring two different pooling sizes, $4 \times 4$ (pooled length and pooled width), and $8 \times 4$ with 4 sampled points in each bin. One of the reasons that we use $8 \times 4$ is that most of the objects like car and bus's length is larger than their width. With all other implementations the same as InfoFocus, Table.~\ref{table:5} presents detection results utilizing the rotated-RoI with different pooling sizes. Compared with the vanilla PointPillars \cite{lang2019PointPillars}, adding the RoIAlign layer with size of $4 \times 4$ increases the mAP performance by 4.4\%. However, InfoFocus still outperforms RoIAlign by 2.5\% with the better information modeling scheme. 

\begin{table}[h!]
\centering
\caption{Comparison with rotated RoIAlign feature extraction results (\%) on the nuScenes validation set}
\begin{tabular}{c c c c c c c c c c c c} 
 \hline
 Method & car & peds. & barri. & traff. & truck & bus & trail. & const. & motor. & bicyc. & mAP \\ [0.5ex] 
 \hline
  RRoI 4x4 & 76.9 & 60.1 & 37.6 & 29.5 & 32.4 & 50.6 & 22.4 & 5.0 & 20.8 & \bf 3.8 & 33.9\\
  RRoI 8x4 & 77.0 & 59.5 & 36.7 & 29.2 & 33.2 & \bf 51.5 & 25.4 & 4.5 & 24.0 & 1.8 & 34.3\\ 
  \hline
  Ours & \bf 77.6 & \bf 61.7 & \bf 43.4 & \bf 33.4 & \bf 35.4 & 50.5 & \bf 25.6 & \bf 8.3 & \bf 25.2 & 2.5 & \bf 36.4 \\ 
 \hline
\end{tabular}
\label{table:5}
\end{table}

\subsection{Real-time Inference Analysis}
As indicated in Table.~\ref{table:4}, our framework takes about 32.9 ms to perform detection on an example of point cloud in the nuScenes, compared with 26.9 ms of the vanilla PointPillars when both are evaluated on a single NVIDIA 1080Ti GPU. In details, the pillar feature extraction time is 12.6 ms, the DCNN costs 1.1 ms, RPN takes 7.3 ms to generate proposals, and the InfoFocus stage takes 11.9 ms. Specifically, the proposal generation for the InfoFocus stage including NMS is 5.1 ms, the PoI feature extraction time is 3.1 ms, and the second stage including three branches takes 0.7 ms. We also note that WYSIWYG \cite{hu2019you} provides the overhead of computing visibility over a 32-beam LiDAR point to be $24.4 \pm 3.5$ ms on average and InfoFocus is faster than WYSIWYG \cite{hu2019you} since we share the similar backbone network. The framework with RROIAlign has an inference time of 32.2 ms. Further, compared with other point-based methods \cite{shi2019pointrcnn,yang2019std}, InfoFocus is considerably faster and conceptually simpler. 

\section{Conclusions}
Non-uniform distribution of point cloud causes varying amount of information at different locations. Therefore, we argue that this imbalance distribution of information may result in degradation on previous 3D voxel-based detectors when modeling 3D objects. To address this issue, we propose a 3D object detection framework with InfoFocus to dynamically conduct information modeling. InfoFocus contain three effective modules including PoI Pooling, the Visibility Attentive Module, and the Adaptive Point-wise Attention. Demonstrated by the comprehensive experiments, our framework achieves the state-of-art performance among all the real-time detectors on the challenging nuScenes dataset.

\noindent\textbf{Acknowledgement}. This work was supported by the Intelligence Advanced Research Projects Activity (IARPA) via DOI/IBC contract numbers D17PC00345 and D17PC00287. The U.S. Government is authorized to reproduce and distribute reprints for Governmental purposes not withstanding any copyright annotation thereon. The authors would like to thank Zuxuan Wu and Xingyi Zhou for proofreading the manuscript.


\clearpage
%
%
\bibliographystyle{splncs04}
\bibliography{egbib}

\begin{thebibliography}{10}
\providecommand{\url}[1]{\texttt{#1}}
\providecommand{\urlprefix}{URL }
\providecommand{\doi}[1]{https://doi.org/#1}

\bibitem{caesar2019nuscenes}
Caesar, H., Bankiti, V., Lang, A.H., Vora, S., Liong, V.E., Xu, Q., Krishnan,
  A., Pan, Y., Baldan, G., Beijbom, O.: nuscenes: A multimodal dataset for
  autonomous driving. arXiv preprint arXiv:1903.11027  (2019)

\bibitem{chen2017multi}
Chen, X., Ma, H., Wan, J., Li, B., Xia, T.: Multi-view 3d object detection
  network for autonomous driving. In: Proceedings of the IEEE Conference on
  Computer Vision and Pattern Recognition. pp. 1907--1915 (2017)

\bibitem{chen2019fast}
Chen, Y., Liu, S., Shen, X., Jia, J.: Fast point r-cnn. In: Proceedings of the
  IEEE International Conference on Computer Vision. pp. 9775--9784 (2019)

\bibitem{Girshick_2015_ICCV}
Girshick, R.: Fast r-cnn. In: The IEEE International Conference on Computer
  Vision (ICCV) (December 2015)

\bibitem{Girshick_2014_CVPR}
Girshick, R., Donahue, J., Darrell, T., Malik, J.: Rich feature hierarchies for
  accurate object detection and semantic segmentation. In: The IEEE Conference
  on Computer Vision and Pattern Recognition (CVPR) (June 2014)

\bibitem{he2017mask}
He, K., Gkioxari, G., Doll{\'a}r, P., Girshick, R.: Mask r-cnn. In: Proceedings
  of the IEEE international conference on computer vision. pp. 2961--2969
  (2017)

\bibitem{hu2019you}
Hu, P., Ziglar, J., Held, D., Ramanan, D.: What you see is what you get:
  Exploiting visibility for 3d object detection. arXiv preprint
  arXiv:1912.04986  (2019)

\bibitem{huang2018improving}
Huang, J., Sivakumar, V., Mnatsakanyan, M., Pang, G.: Improving rotated text
  detection with rotation region proposal networks. arXiv preprint
  arXiv:1811.07031  (2018)

\bibitem{kingma2014adam}
Kingma, D.P., Ba, J.: Adam: A method for stochastic optimization. arXiv
  preprint arXiv:1412.6980  (2014)

\bibitem{ku2018joint}
Ku, J., Mozifian, M., Lee, J., Harakeh, A., Waslander, S.L.: Joint 3d proposal
  generation and object detection from view aggregation. In: 2018 IEEE/RSJ
  International Conference on Intelligent Robots and Systems (IROS). pp.~1--8.
  IEEE (2018)

\bibitem{Lan_2019_CVPR}
Lan, S., Yu, R., Yu, G., Davis, L.S.: Modeling local geometric structure of 3d
  point clouds using geo-cnn. In: The IEEE Conference on Computer Vision and
  Pattern Recognition (CVPR) (June 2019)

\bibitem{lang2019PointPillars}
Lang, A.H., Vora, S., Caesar, H., Zhou, L., Yang, J., Beijbom, O.:
  Pointpillars: Fast encoders for object detection from point clouds. In:
  Proceedings of the IEEE Conference on Computer Vision and Pattern
  Recognition. pp. 12697--12705 (2019)

\bibitem{liang2019multi}
Liang, M., Yang, B., Chen, Y., Hu, R., Urtasun, R.: Multi-task multi-sensor
  fusion for 3d object detection. In: Proceedings of the IEEE Conference on
  Computer Vision and Pattern Recognition. pp. 7345--7353 (2019)

\bibitem{Liang_2018_ECCV}
Liang, M., Yang, B., Wang, S., Urtasun, R.: Deep continuous fusion for
  multi-sensor 3d object detection. In: The European Conference on Computer
  Vision (ECCV) (September 2018)

\bibitem{lin2017focal}
Lin, T.Y., Goyal, P., Girshick, R., He, K., Doll{\'a}r, P.: Focal loss for
  dense object detection. In: Proceedings of the IEEE international conference
  on computer vision. pp. 2980--2988 (2017)

\bibitem{liu2019point}
Liu, Z., Tang, H., Lin, Y., Han, S.: Point-voxel cnn for efficient 3d deep
  learning. In: Advances in Neural Information Processing Systems. pp. 963--973
  (2019)

\bibitem{qi2018frustum}
Qi, C.R., Liu, W., Wu, C., Su, H., Guibas, L.J.: Frustum pointnets for 3d
  object detection from rgb-d data. In: Proceedings of the IEEE Conference on
  Computer Vision and Pattern Recognition. pp. 918--927 (2018)

\bibitem{qi2017pointnet}
Qi, C.R., Su, H., Mo, K., Guibas, L.J.: Pointnet: Deep learning on point sets
  for 3d classification and segmentation. In: Proceedings of the IEEE
  conference on computer vision and pattern recognition. pp. 652--660 (2017)

\bibitem{qi2017pointnet++}
Qi, C.R., Yi, L., Su, H., Guibas, L.J.: Pointnet++: Deep hierarchical feature
  learning on point sets in a metric space. In: Advances in neural information
  processing systems. pp. 5099--5108 (2017)

\bibitem{ren2015faster}
Ren, S., He, K., Girshick, R., Sun, J.: Faster r-cnn: Towards real-time object
  detection with region proposal networks. In: Advances in neural information
  processing systems. pp. 91--99 (2015)

\bibitem{shi2019pointrcnn}
Shi, S., Wang, X., Li, H.: Pointrcnn: 3d object proposal generation and
  detection from point cloud. In: Proceedings of the IEEE Conference on
  Computer Vision and Pattern Recognition. pp. 770--779 (2019)

\bibitem{simonelli2019disentangling}
Simonelli, A., Bulo, S.R., Porzi, L., L{\'o}pez-Antequera, M., Kontschieder,
  P.: Disentangling monocular 3d object detection. In: Proceedings of the IEEE
  International Conference on Computer Vision. pp. 1991--1999 (2019)

\bibitem{smith2018disciplined}
Smith, L.N.: A disciplined approach to neural network hyper-parameters: Part
  1--learning rate, batch size, momentum, and weight decay. arXiv preprint
  arXiv:1803.09820  (2018)

\bibitem{vora2019pointpainting}
Vora, S., Lang, A.H., Helou, B., Beijbom, O.: Pointpainting: Sequential fusion
  for 3d object detection. arXiv preprint arXiv:1911.10150  (2019)

\bibitem{wang2019frustum}
Wang, Z., Jia, K.: Frustum convnet: Sliding frustums to aggregate local
  point-wise features for amodal 3d object detection. arXiv preprint
  arXiv:1903.01864  (2019)

\bibitem{yan2018second}
Yan, Y., Mao, Y., Li, B.: Second: Sparsely embedded convolutional detection.
  Sensors  \textbf{18}(10), ~3337 (2018)

\bibitem{yang2019std}
Yang, Z., Sun, Y., Liu, S., Shen, X., Jia, J.: Std: Sparse-to-dense 3d object
  detector for point cloud. In: Proceedings of the IEEE International
  Conference on Computer Vision. pp. 1951--1960 (2019)

\bibitem{ye2020sarpnet}
Ye, Y., Chen, H., Zhang, C., Hao, X., Zhang, Z.: Sarpnet: Shape attention
  regional proposal network for lidar-based 3d object detection. Neurocomputing
   \textbf{379},  53--63 (2020)

\bibitem{zhou2019objects}
Zhou, X., Wang, D., Kr{\"a}henb{\"u}hl, P.: Objects as points. arXiv preprint
  arXiv:1904.07850  (2019)

\bibitem{zhou2018voxelnet}
Zhou, Y., Tuzel, O.: Voxelnet: End-to-end learning for point cloud based 3d
  object detection. In: Proceedings of the IEEE Conference on Computer Vision
  and Pattern Recognition. pp. 4490--4499 (2018)

\end{thebibliography}
\end{document}